\documentclass[11pt,a4paper]{article}
\usepackage[hyperref]{acl2021}
\usepackage{times}
\usepackage{latexsym}

\usepackage{arydshln}
\usepackage{microtype}

\aclfinalcopy 
\def\aclpaperid{2354} 

\usepackage{amsfonts}       
\usepackage{amsmath}
\usepackage{amssymb}
\usepackage{graphicx}
\usepackage{enumitem}
\usepackage{breakcites}
\usepackage{multicol}
\usepackage{multirow}

\usepackage{url}

\usepackage{arydshln}
\usepackage{algorithm}          
\usepackage{algpseudocode}      
\usepackage{amsmath}            
\usepackage{amssymb}            
\usepackage{float}

\usepackage{xspace}
\newcommand*{\eg}{e.g.\@\xspace}
\newcommand*{\ie}{i.e.\@\xspace}

\newcommand{\augnlg}{\textsc{AugNLG}}
\newcommand{\fewshotwoz}{\textsc{FewShotWOZ}}
\newcommand{\fewshotsgd}{\textsc{FewShotSGD}}

\title{\augnlg: {F}ew-shot {N}atural {L}anguage {G}eneration using {S}elf-trained {D}ata {A}ugmentation}

\author{Xinnuo Xu$^1$, Guoyin Wang$^2$, Young-Bum Kim$^2$, Sungjin Lee$^2$ \\
  $^1$The Interaction Lab, Heriot-Watt University, Edinburgh\\
  $^2$Amazon Alexa AI, Seattle, WA, USA \\
  {\tt xx6@hw.ac.uk, guoyiwan, youngbum, sungjinl@amazon.com} \\}

\date{}

\begin{document}
\maketitle
\begin{abstract}

Natural Language Generation (NLG) is a key component in a task-oriented dialogue system, which converts the structured meaning representation (MR) to the natural language.
For large-scale conversational systems, where it is common to have over hundreds of intents and thousands of slots, neither template-based approaches nor model-based approaches are scalable.
Recently, neural NLGs started leveraging transfer learning and showed promising results in few-shot settings.
This paper proposes {\augnlg}, a novel data augmentation approach that combines a self-trained neural retrieval model with a few-shot learned NLU model, to automatically create MR-to-Text data from open-domain texts.
The proposed system mostly outperforms the state-of-the-art methods on the {\fewshotwoz} data in both BLEU and Slot Error Rate. 
We further confirm improved results on the {\fewshotsgd} data and provide comprehensive analysis results on key components of our system. Our code and data are available at \url{https://github.com/XinnuoXu/AugNLG}.

\end{abstract}

\section{Introduction}
\begin{figure}[t]
	\includegraphics[width=1.0\columnwidth]{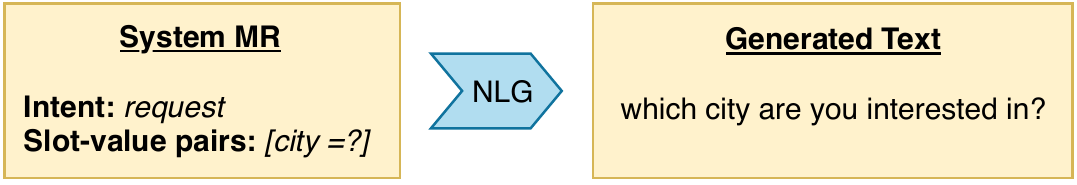}
	\caption{An example of NLG task. The model takes in the system MR, which consists of an intent with slot value pairs, and outputs text in natural language.}
	\label{fig:ds}
\end{figure}

Large-scale conversational systems provide a natural interface to achieve various daily-life tasks. Natural Language Generation (NLG) is a key component in such a system to convert the structured meaning representation (MR) to the natural language, as shown in \autoref{fig:ds}. In task-oriented dialogue systems, NLG is typically accomplished by filling out a basic set of developer-provided templates, leading to a conversational system generating unnatural, robotic responses. In order to make the system sound more human-like, model-based NLG approaches, in particular neural models, have recently been gaining an increasing traction \cite{gao-etal-2018-neural, wen-etal-2015-semantically}. 
However, neither the template-based approaches nor the model-based approaches are sufficiently scalable for large-scale conversational systems, where it is common to have over hundreds of intents and thousands of slots.

With the rise of neural transfer learning for NLP using pretrained LMs, recently, neural NLGs started to leverage transfer learning and showed some promising results \cite{radford2019language, brown2020language, dai2019transformer, edunov2019pre}. In particular, \citet{peng-etal-2020-shot} proposed {\fewshotwoz}, the first NLG benchmark test in few-shot learning settings, and achieved a SOTA performance 
by leveraging existing MR-to-Text data sets via task-specific continued pre-training. Despite the improved result, their approach leaves little room for further improvements as MR-to-Text data are expensive to obtain for new domains, practically circling back to the same scalability problem after exhausting the existing data. 

In order to go beyond this restriction, this paper proposes {\augnlg}, a novel data augmentation approach, that automatically creates MR-to-Text data from open-domain texts by combining a self-trained neural retrieval model with a few-shot learned NLU model. Since our data augmentation approach is orthogonal to the prior transfer learning approaches, one can use our approach in conjunction with other approaches. In experiments, we empirically show that {\augnlg} mostly boosts the performance of both the fine-tuned GPT-2 (FT-GPT) \cite{radford2019language} and SC-GPT \cite{peng-etal-2020-shot}, the continued pretraining approach with existing MR-to-Text data, on the {\fewshotwoz} task. Furthermore, we construct another few-shot learning testbed, {\fewshotsgd}, out of the Schema-Guided Dialogue (SGD) corpus \cite{rastogi2020schema} and confirm improved results by applying {\augnlg} to the FT-GPT.~\footnote{Since SGD accounts for a large portion of the existing MR-to-Text data that SC-GPT utilized in training, we could not apply {\augnlg} to SC-GPT for the {\fewshotsgd} task.} Finally, we provide comprehensive analysis results on the key components of our system to gain detailed insights into the relationship between component-wise behavior and various parameters.

\begin{figure}[tb]
	\centering
	\includegraphics[width=1.0\columnwidth]{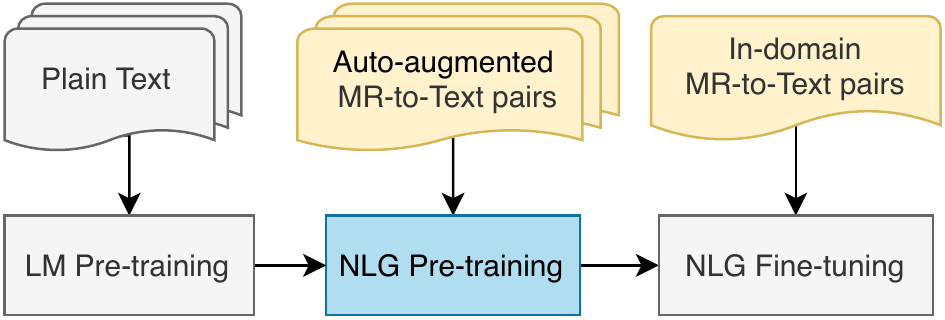}
	\caption{The training procedure for {\augnlg}. }
	\label{fig:model_training}
\end{figure}

\begin{figure*}[tb]
	\centering
	\includegraphics[width=0.9\textwidth]{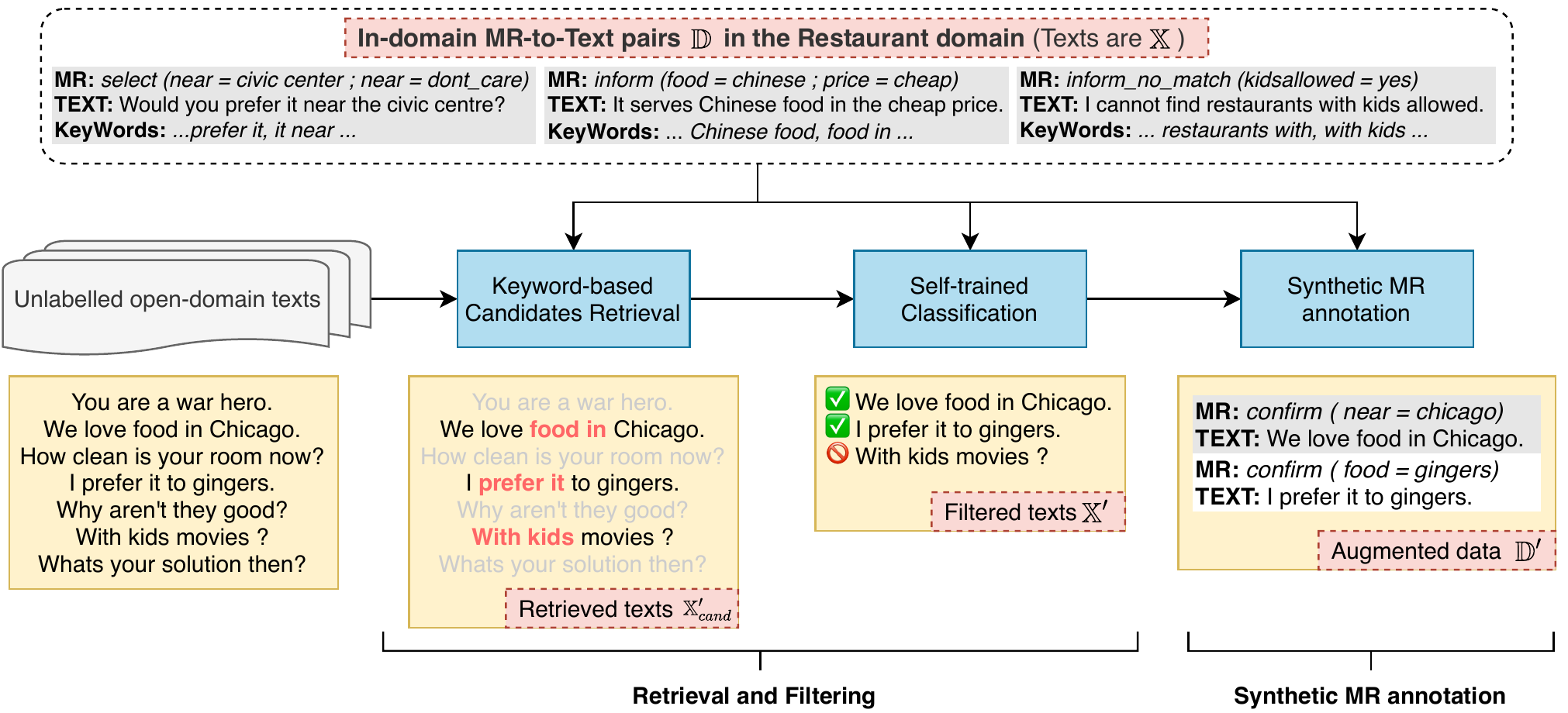}
	\caption{The overall pipeline for MR-to-Text data augmentation.}
	\label{fig:data_aug}
\end{figure*}

\section{Related Work}
\textbf{NLG for Dialogue Response Generation} 
There has been a body of work on neural NLG models, adopting various architectures, such as RNNs \cite{wen-etal-2015-semantically}, attention RNNs \cite{dusek-jurcicek-2016-sequence}, SC-LSTM \cite{wen-etal-2016-multi}, T2G2 \cite{kale-rastogi-2020-template}, AdapterCL \cite{madotto2020continual} and associated variants \cite{tran2017natural, tran2017neural}. Despite the improved flexibility and naturalness over template-based methods, neural approaches require large amounts of annotated data to reach good performance.

\noindent \textbf{Data Augmentation} 
Data augmentation has been widely applied to a variety of NLP tasks, including sentence classification \cite{xie2020unsupervised}, natural language inference \cite{NEURIPS2019_671f0311} and spoken language understanding \cite{li2019insufficient, quan2019effective, zhao2019data}. Prior approaches for text data utilized back-translation \cite{sennrich2016improving, edunov2018understanding}, c-BERT word replacement \cite{jiao-etal-2020-tinybert}, mixed labels and representations  
\cite{guo2019augmenting, chen2020mixtext} and paraphrase data \cite{gao2020paraphrase}. However, the range of augmented data will be inherently limited, particularly in few-shot learning settings due to the nature of prior approaches, which only leverages in-domain data. In contrast, we take a rarely explored approach, tapping into a wealth of open-domain text that covers almost all topics. Recently, \citet{du-etal-2021-self} proposed a self-training method to augment data for NLU tasks by retrieving sentences from data crawled on the web. However, their method cannot be directly applied to the NLG problem since it does not yield MR annotations. Our approach, in contrast, generates MR-to-Text data by jointly employing a self-trained neural retrieval model with a few-shot learned NLU model.

\section{Few-shot Transfer Learning for NLG}

The goal of NLG is to translate an MR $\mathcal{A}$ into its natural language response $x = \left[x^1, \ldots, x^T \right]$, where ${x^i}$ is the $i$th token in the sequence $x$ and $T$ is the sequence length. $\mathcal{A}$ is defined as the combination of intent $\mathcal{I}$ and slot-value pairs $ \{ ( s_i, v_i ) \}_{i=1}^{P}$:
\begin{equation}\label{Eq:1}
	\mathcal{A} = \{ \mathcal{I}, (s_1, v_1), \dots, (s_P, v_P)\},
\end{equation}

where the intent stands for the illocutionary type of the system action 
while slot-value pairs indicate category names and their values to embed in the utterance. For example, in the MR, \textit{inform (food = chinese ; price = cheap)}, \textit{inform} is the intent, \textit{food} and \textit{price} are two slot keys and \textit{chinese} and \textit{cheap} are the corresponding slot values.


Given in-domain MR-to-Text data $\mathbb{D}=\{(\mathcal{A}_n, x_n)\}_{n=1}^{N}$ for training, where $N$ is the number of examples, a statistical neural language model parameterized by $\theta$ is adopted to characterize the conditional probability $p_{\theta}(x|\mathcal{A})$. 
By adopting the chain rule on auto-regressive generation, the joint probability of $x$ conditioned on $\mathcal{A}$ is decomposed as $\prod_{t=1}^{T} p_{\theta}(x^t|x^{<t}, \mathcal{A})$. The training process, i.e. the learning of $\theta$, is then defined as maximizing the log-likelihood of the conditional probabilities over the entire training dataset:
\begin{equation*}\label{Eq:2}
	\mathcal{L}_{\theta}(\mathbb{D})=\sum_{n=1}^{\left\lvert \mathbb{D} \right\rvert} \log p_{\theta}(x_n|\mathcal{A}_n) .
\end{equation*}

In the few-shot learning setup, the number of training examples $N$ is extremely small (\eg $\leq 50$), which easily leads to non-fluent generated sentences with many grammar mistakes or missing pieces of information. 
In order to combat the data sparseness problem, inspired by prior transfer learning approaches, 
we introduce a three-step pipeline to gradually evolve a general large-scale language model to a domain-specific NLG model (shown in \autoref{fig:model_training}): (1) pre-training a base language model with massive amounts of text, (2) NLG-specific continued pre-training with auto-augmented MR-to-Text data, and (3) final fine-tuning with the limited in-domain MR-to-Text ground-truth data. 

Specifically, in Step (1), we adopt GPT-2 \cite{radford2019language} as our base language model since GPT-2 has demonstrated a remarkable performance on auto-regressive text generation tasks, which is close to MR-to-Text generation, in a variety of domains.
However, GPT-2 is pre-trained on 
OpenWebText and the language style and topics thereof are quite different from those of daily conversations in a target domain.
Furthermore, the generation task in NLG is conditioned on the input MR, as opposed to the unconditioned generation of the underlying GPT-2 pre-training task. 
Thus, to bring the model a step closer to the final NLG model in the target domain, in Step (2), we continuously pre-train the GPT-2 model on an automatically constructed set of augmented MR-to-Text pairs $\mathbb{D'}=\{(\mathcal{A}_m, x_m)\}_{m=1}^{M}$, where $M$ is the number of augmented examples, which is much larger than the amount of in-domain ground-truth data. Data augmentation is achieved by retrieving a large amount of relevant text from Reddit \cite{Henderson2019} with a self-trained neural retrieval model and then synthesizing MRs with a few-shot learned NLU model. The details of data augmentation is described in \autoref{sec:augmentation}. 
Finally, in Step (3), we fine-tune the NLG model on a limited amount of in-domain ground-truth MR-to-Text pairs $\mathbb{D}$ for a final adaptation.
\vspace{-5pt}
\section{Data Augmentation} \label{sec:augmentation}
\vspace{-5pt}

The data augmentation procedure aims to construct a large amount of MR-to-Text pairs $\mathbb{D}'$ from open-domain texts that are relevant to the in-domain ground-truth MR-to-Text pairs $\mathbb{D}$. 
The augmentation process consists of two stages: (1) retrieving  keyword-matching utterances and filtering out domain-irrelevant instances, (2) generating synthetic MR annotations. \autoref{fig:data_aug} illustrates the overall pipeline with some examples. For further analysis and studies, we release the data from all intermediate steps for each domain at \url{https://github.com/XinnuoXu/AugNLG/tree/master/augmented_data}.

\subsection{Retrieval and Filtering}\label{subsec:retrieval}
The utterance retrieval and filtering procedure consists of three steps: (1) keyword extraction that collects n-gram keywords from all in-domain utterances $\mathbb{X}=\left \{ x_n \right \}_{n=1}^{N}$; (2) keyword-based retrieval that searches the open-domain texts for utterances that match any keywords extracted in the previous step, yielding a set of utterances $\mathbb{X}'_{cand}$; (3) self-trained neural classifier that filters out some retrieved utterances that are semantically irrelevant to the target domain.
After the filtering, we form an augmented set of utterances $\mathbb{X}'$ with the unfiltered utterances.

\paragraph{Keywords Extraction.} To efficiently extract keywords, 
we first gather all n-gram phrases that appear in $\mathbb{X}$. 
Since some phrases are too general to be effective, \eg ``\textit{I cannot}'', ``\textit{is your}'',
we use TF-IDF scores to measure the specificity of a phrase (see Appendix~\ref{sec:sup-tf-idf} for more detail).
We first rank the collected n-grams according to their TF-IDF scores and filter out those n-gram phrases with relatively low TF-IDF score. 

\paragraph{Keyword-based Retrieval.} Having extracted the keywords, we retrieve utterances from the open-domain utterance pool that contains at least one extracted keyword in it. The aim of this step is to source a large amount of domain-relevant utterances $\mathbb{X}'_{cand}$ based on the surface-level overlap.


\paragraph{Self-trained Neural Filtering.} Although the keyword-based retrieval is efficient, the retrieved utterances $\mathbb{X}'_{cand}$ can be quite noisy since an n-gram keyword only matches some part of the utterance, failing to detect the existence of irrelevant pieces in other parts.
For example, in \autoref{fig:data_aug}, even though the utterance ``\textit{With kids movies?}'' contains the keyword ``\textit{with kids}'', it is irrelevant to  the target domain \emph{Restaurant} given the word \emph{movies}. 
Thus, we introduce a self-trained neural classifier 
to filter out domain-irrelevant utterances from $\mathbb{X}'_{cand}$ by considering the semantic representation of an entire utterance and yield a domain-relevant set $\mathbb{X}'$. 


\begin{algorithm}[tb]
    \begin{algorithmic}[1]
        \caption{Self-trained Neural Filtering} \label{algorithm:1}
		\Require In-domain utterances $\mathbb{X}$ in the target domain; Retrieved utterances $\mathbb{X}'_{cand}$
		\State $\mathcal{U}^{+} \gets$ Positive examples $\mathbb{X}$
        \State $\mathcal{U}^{-} \gets $ Randomly selected negative examples
		\State $c^0 \gets $ Train($\mathcal{U}^{+}$, $\mathcal{U}^{-}$)
		\State $L \gets$ Maximum number of iterations
		\State $l = 1$; $\mathcal{E}^{+}_0=\mathcal{U}^{+}$; $\mathcal{E}^{-}_0=\mathcal{U}^{-}$
		\While{$l \leq L$}
			\State $\mathcal{E}^{+}_l \gets \{x'$ if Predict($\mathbb{X}'_{cand}$, $c^{l-1}$) $\geq \sigma^{+} \}$
			\State $\mathcal{E}^{-}_l \gets \{x'$ if Predict($\mathbb{X}'_{cand}$, $c^{l-1}$) $\leq \sigma^{-} \}$
			\State $\mathcal{E}^{+}_l \gets \mathcal{E}^{+}_l + \mathcal{U}^{+}$
			\If{$\left | \mathcal{E}^{+}_l \right | - \left | \mathcal{E}^{+}_{l-1} \right | \leq \delta$}
			\State Converged; Break
			\EndIf
			\State $c^l \gets $ Train($\mathcal{E}^{+}_l$, $\mathcal{E}^{-}_l$)
			\State $l \gets l+1$
        \EndWhile
    \State $\mathbb{X}' \gets \{x'$ if Predict($\mathbb{X}'_{cand}$, $c^{l}$) $\geq \sigma \}$ 
    \end{algorithmic}
\end{algorithm}

The algorithm of the self-training and filtering process is listed in Algorithm~\ref{algorithm:1}.
We adopt a BERT \cite{devlin-etal-2019-bert} model with a binary classification layer atop as the base model and then train the classifier with in-domain utterances $\mathbb{X}$ and randomly selected open-domain utterances\footnote{
    All utterances in $\mathbb{X}'_{cand}$ are excluded from the open-domain utterance pool. To balance the precision and recall, we control the size of the initial negative set such that $ \left | \mathcal{U}^{-}  \right | = \lambda_1 \cdot \left | \mathcal{U}^{+}  \right | $, where $\lambda_1=10$.
    }
, serving as positive and negative examples ($\mathcal{U}^{+}$ and $\mathcal{U}^{-}$), respectively. 
After that, the self-training and filtering cycle starts. 
At each iteration, we make predictions on the utterances in $\mathbb{X}'_{cand}$ with the classifier trained in the previous iteration. 
All utterances with a score over the threshold $\sigma^{+}$, together with the in-domain utterances $\mathbb{X}$, are then taken as a new set of positive examples $\mathcal{E}^{+}$, whereas all utterances with a score less than the threshold $\sigma^{-}$ are collected as a new set of negative examples $\mathcal{E}^{-}$.\footnote{
    To guarantee the precision of the positive examples, we use $\sigma^{+}=0.99$ and $\sigma^{-}= 0.5$. 
    Also, we sub-sample negative examples such that $ \left | \mathcal{E}^{-}  \right | = \lambda_2 \cdot \left | \mathcal{E}^{+}  \right | $, where $\lambda_2=5$.} 
The self-training loop terminates if either the increment of positive examples at the last iteration is less than the threshold $\delta$ or the iterations is over the pre-defined maximum number of iterations. Otherwise, a new classifier is trained on $\mathcal{E}^{+}$ and $\mathcal{E}^{-}$ and the algorithm keeps going on the loop.
Once the loop terminated, we label all utterances in $\mathbb{X}'_{cand}$ with the classifier from the last iteration.
Finally, we build a domain-relevant set of augmented utterances $\mathbb{X}'$ by taking all utterances with a score over the threshold $\sigma$.\footnote{To harvest a large amount of utterances, we set the threshold $\sigma$ to 0.5. } 

\subsection{Synthetic MR Annotation}\label{subsec:annotation}
Having built the domain-relevant set of augmented utterances $\mathbb{X}'$, we now proceed to synthesize MR labels to produce a complete MR-to-Text dataset $\mathbb{D}'$.
To this end, we build a few-shot NLU model by fine-tuning a BERT model with in-domain ground-truth data. To put the data in the right format for the NLU task, we take MRs and utterances as labels and model inputs, respectively. 
Each token is annotated with the slot name if it is a part of the associated slot value and 
the final hidden state of the special token [CLS] is used to predict the intent (see \autoref{fig:nlu} in Appendix~\ref{sec:sup-nlu}). 
Finally, we generate an MR-to-Text dataset $\mathbb{D}'$ by concatenating the utterances in $\mathbb{X}'$ with the synthetic MR labels predicted by the few-shot NLU model.

\section{Experimental Setup}
\subsection{Dataset}

\begin{table}[t]
	\small\centering
	\begin{tabular}{l|cc}
	\hline
	Statistics	& {\bf -WOZ} & {\bf -SGD}  \\ \hline
	\# Domains	& 7 & 16 \\ 
	Avg. \# Intents	& 8.14 & 6.44 \\
	Avg. \# Slots	& 16.2 & 11.3 \\
	Avg. \# Delex MRs in Training	& 50 & 33 \\
	Avg. \# Delex MRs in Testing	& 473 & 31 \\
	Avg. \# Training Instances	& 50 & 35 \\
	Avg. \# Test Instances	& 473 & 5618 \\
	Avg. \# Test Instances per MR	& 1.14 & 472.9 \\
	Avg. \# Test Novelty uni-gram (\%)	& 12.97 & 23.90 \\
	Avg. \# Test Novelty bi-gram(\%)	& 44.42 & 65.29 \\
	Avg. \# Test Novelty tri-gram(\%)	& 68.20 & 84.44 \\
	Avg. \# Test Novelty four-gram(\%)	& 82.70 & 92.75 \\ \hline
	Avg. \# Keywords (K) & 0.20 & 0.12 \\
	Avg. \# Retrieved Utterances (K) & 854.8 & 731.3 \\
	Avg. \# Augmented Pairs (K) & 34.0 & 25.6 \\
	Avg. \# Delex. MRs in Aug. Pairs (K) & 2.12 & 0.57 \\
	\hline
	\end{tabular}
	\caption{Comparison of {\fewshotwoz} and {\fewshotsgd}. The bottom section shows the statistics for augmented data. The unit for all statistics in the bottom section is thousand(K). 
	\label{table:data_fewshot}}
\end{table}

\paragraph{Fewshot NLG Data}
{\fewshotwoz} is a few-shot NLG benchmark, built upon RNNLG and MultiWOZ \cite{budzianowski2018multiwoz}.
In each domain, MR-to-Text pairs are grouped according to their delexicalized MRs (\ie slot values being masked) and a training set is created by taking a pair each from 50 random groups and then the rest are taken as the test set. 
We also construct a new dataset {\fewshotsgd} by applying the same preparation steps to the SGD corpus.
The comparison of {\fewshotwoz} and {\fewshotsgd} is presented in the top section in \autoref{table:data_fewshot}. 
Comparing to {\fewshotwoz}, {\fewshotsgd} has (1) more domains, (2) less intents, slots and delexicalized MRs\footnote{
    Note that, the average number of delexicalized MRs in the training set is 33,
    which means the number of training examples in some domains are less than 50.}
(3) more testing examples for each delexicalized MR, 
(4) more novel n-grams\footnote{The novelty is calculated by dividing the number of n-grams in the test set that does not appear in the training set by the number of n-grams in the test set.} in test utterances.


\begin{table*}[t]
	\small\centering
	\begin{tabular}{l|p{0.5cm}p{0.5cm}|p{0.5cm}p{0.5cm}|p{0.5cm}p{0.5cm}|p{0.5cm}p{0.5cm}|p{0.5cm}p{0.5cm}|p{0.5cm}p{0.5cm}|p{0.5cm}p{0.5cm}}
	\hline
	{\bf Model}  & \multicolumn{2}{c|}{\bf Restaurant}  & \multicolumn{2}{c|}{\bf Laptop} & \multicolumn{2}{c|}{\bf Hotel}  & \multicolumn{2}{c|}{\bf TV} & \multicolumn{2}{c|}{\bf Attraction} & \multicolumn{2}{c|}{\bf Train} & \multicolumn{2}{c}{\bf Taxi} \\ 
				 & BLEU & ERR  & BLEU & ERR  & BLEU & ERR  & BLEU & ERR  & BLEU & ERR  & BLEU & ERR  & BLEU & ERR\\ \hline\hline
	FT-GPT        & 28.15 & 15.87 & 28.83 & 11.82 & 36.51 & 14.29 & 33.73 & 9.28 & 17.45 & 22.83 & 13.06 & 25.59 & 14.84 & 28.57 \\ 
	{\augnlg}-FT      & 32.16 & 4.79 & 33.64 & 5.14 & 36.99 & 9.89 & 34.80 & 6.92 & 20.61 & 13.58  & 14.95 & 10.64 & 16.70 & 10.71 \\ \hdashline
	SC-GPT    & 30.48 & 6.89 & 33.51 & 5.38 & {\bf 38.30} & 8.24 & 33.82 & 7.32 & 22.24 & 16.62 & {\bf 17.06} & 8.82 & {\bf 19.21} & 4.76 \\
	{\augnlg}-SC & {\bf 34.20} & {\bf 2.99} & {\bf 34.32} & {\bf 2.83} & 34.96 & {\bf 6.59} &  {\bf 34.99} & {\bf 5.53} & {\bf22.50} & {\bf 10.40} & 16.35 & {\bf 6.13} & 17.81 & {\bf 3.57} \\
	\hline
	\end{tabular}
	\caption{Evaluation results on {\fewshotwoz} (BLEU$\uparrow$, ERR$\downarrow$). Note that, the SC-GPT model reported here was pre-trained and fine-tuned using the code and only the SGD data shared by the original authors \protect\footnotemark. 
	\label{table:fewshotwoz_res}}
\end{table*}
\footnotetext{\url{https://github.com/pengbaolin/SC-GPT}.}

\begin{table*}[t]
	\small\centering
	\begin{tabular}{l|cccccccc}
	\hline
	{\bf Model}        & {\bf Restaurants} & {\bf Hotels} & {\bf Flights} & {\bf Calendar} & {\bf Banks} & {\bf Weather} & {\bf Buses} & {\bf Events} \\ \hdashline
	FT-GPT        & 08.98 & 08.84 & 12.18 & 05.27 & 06.09 & 10.52 & 07.77 & 09.17 \\ 
	{\augnlg}-FT      & {\bf 17.83} & {\bf 17.23} & {\bf 17.58} & {\bf 10.45} & {\bf 08.94} & {\bf 13.75} & {\bf 14.26} & {\bf 18.68} \\
	\hline\hline
	{\bf Model}        & {\bf Homes} & {\bf Media} & {\bf Movies} & {\bf Music} & {\bf Rentalcars} & {\bf Ridesharing} & {\bf Services} & {\bf Travel} \\ \hdashline
	FT-GPT        & 03.75 & 03.17 & 10.05 & 05.79 & 06.79 & 13.87 & 09.79 & 02.08 \\ 
	{\augnlg}-FT      & {\bf 12.27} & {\bf 08.62} & {\bf 11.96} & {\bf 12.76} & {\bf 13.32} & {\bf 15.54} & {\bf 16.82} & {\bf 14.35} \\
	\hline
	\end{tabular}
	\caption{Evaluation results in BLEU on {\fewshotsgd}. 
    \label{table:fewshotsgd_res}}
\end{table*}

\paragraph{Augmented Data}
Since Reddit has shown to provide natural conversational English data,
we adopt Reddit \cite{Henderson2019} as the open-domain utterance pool after filtering for utterances of length between 2 and 40, totalling about 0.7B utterances. 
The average number of extracted keywords, retrieved utterances, final augmented MR-to-Text pairs and delexicalized MRs 
over all domains in {\fewshotwoz} and {\fewshotsgd} are shown in the bottom section of \autoref{table:data_fewshot}. The detailed breakdowns of each domain are listed in \autoref{table:augment_data_woz} and \autoref{table:augment_data_sgd} in Appendix~\ref{sec:sup-augdata-statistic}.

\subsection{Evaluation Metrics}
Following \citet{wen-etal-2015-semantically} and \citet{peng-etal-2020-shot}, we use BLEU score and Slot Error Rate (ERR) for automatic evaluation. 
BLEU score measures the surface-level similarity between generated responses and human-authored references. 
Whereas, ERR 
measures the semantic alignment in terms of slot-value insertion and omission.
Specifically, $\textup{ERR}=(p + q)/M$, where $M$ is the total number of slots in the MR and $p$, $q$ are the number of missing and redundant slots in the surface realisation. Since the SGD dataset does not provide enough information to compute ERR, we report ERR only on {\fewshotwoz}.


\subsection{Systems}
We apply our data augmentation approach {\augnlg} to two baseline systems,
\begin{itemize}
    \item {\bf FT-GPT} GPT-2 is directly fine-tuned on the in-domain ground-truth MR-to-Text data. We introduce {\bf \augnlg{}-FT}, which further pre-trains GPT-2 on the augmented MR-to-Text data and performs a final fine-tuning on the in-domain data. 
    \item {\bf SC-GPT} \cite{peng-etal-2020-shot} further pre-trains GPT-2 on existing MR-to-Text data borrowed from other NLG corpora and fine-tunes on the in-domain data. We introduce {\bf \augnlg{}-SC}, which pre-trains GPT-2 on both existing MR-to-Text data and automatically augmented data, and finally fine-tunes on the in-domain data.     
\end{itemize}


\section{Results}
\paragraph{\fewshotwoz{}} \autoref{table:fewshotwoz_res} reports the results on {\fewshotwoz{}}. 
{\augnlg}-FT substantially outperforms FT-GPT across all domains in both BLEU and ERR. 
Similarly, {\augnlg}-SC performs better than SC-GPT and achieves the state-of-the-art performance in most domains.
Remarkably, {\augnlg}-FT achieves a competitive performance with SC-GPT in many domains without leveraging any existing MR-to-Text data. It even outperforms SC-GPT in ``\textit{TV}'' and ``\textit{Attraction}'' domain in both BLEU and ERR.

\paragraph{\fewshotsgd{}} \autoref{table:fewshotsgd_res} shows the results in {\fewshotsgd}. 
Due to the higher novelty of the test examples and the smaller amount of training examples (see {\em Avg. \# Test Novelty n-gram} and {\em \# Training Instances} in \autoref{table:data_fewshot}), FT-GPT performs worse than on {\fewshotwoz}.
This indicates that the few-shot settings on {\fewshotsgd} are even more challenging. But {\augnlg}-FT managed to outperform FT-GPT by a large margin via the continued pre-training on the augmented examples. 

\begin{table*}
\centering\small
\begin{tabular}{l|l}
\hline
\multicolumn{2}{c}{\textbf{Domain: \textit{Restaurant }}} \\ \hdashline
Input MR    & \textit{inform(name=marlowe; goodformeal=dinner; area=mission bay)}\\
Reference   & marlowe serves dinner in the mission bay area.\\
FT-GPT   & there is a restaurant marlowe in the mission bay area \textcolor{blue}{called dinner},~it is good for dinner, \textcolor{blue}{a good area}.\\
SC-GPT  & marlowe is a good restaurant in mission bay that serves dinner.\\
\augnlg-FT    & marlowe is a good restaurant that serves good dinner in mission bay.\\
\augnlg-SC & marlowe is good for dinner in mission bay.\\
\hline

\multicolumn{2}{c}{\textbf{Domain:\textit{ Laptop }}}\\ \hdashline
Input MR    & \textit{inform(name=tecra proteus 23;type=laptop;battery=9 hour;memory=4 gb;isforbusiness=false)}\\
Reference   & the tecra proteus 23 laptop has a 9 hour battery life with 4 gb memory but is not for business computing.\\
FT-GPT   & tecra proteus 23 laptop with 9 hour battery and 4 gb memory. \textbf{\textcolor{red}{\% miss}\textit{\textcolor{red}{~\{isforbusiness=false\}}}} \\
SC-GPT  & tecra proteus 23 laptop with 4 gb memory and 9 hour battery. \textbf{\textcolor{red}{\% miss}\textit{\textcolor{red}{~\{isforbusiness=false\}}}} \\
\augnlg-FT & the tecra proteus 23 laptop has a 9 hour battery, 4 gb of memory, and is not for business computing.\\
\augnlg-SC  & tecra proteus 23 laptop - 9 hour battery - 4 gb memory - not used for business computing.\\
\hline

\multicolumn{2}{c}{\textbf{Domain:\textit{~TV}}}\\ \hdashline
Input MR    & \textit{inform(name=pontus 45; type=tv;ecorating=a+; screensizerange=medium; hdmiport=2)}\\
Reference   & the pontus 45 tv has a a+ eco rating and a medium screen size and 2 hdmi ports. \\
FT-GPT   & the pontus 45 tv has 2 hdmi ports, a medium screen size , \textcolor{blue}{and 2 screensize}. \textit{\textbf{\textcolor{red}{\% miss \{ecorating=a+\}}}}\\
SC-GPT  & the pontus 45 tv has a medium screen size and 2 hdmi ports. \textit{\textbf{\textcolor{red}{\% miss \{ecorating=a+\}}}} \\
\augnlg-FT & the pontus 45 tv has a+ eco rating, 2 hdmi ports, and a medium screen size.\\
\augnlg-SC & the pontus 45 tv has a medium screen size of 2 hdmi ports and a+ eco rating.\\
\hline

\multicolumn{2}{c}{\textbf{Domain:~\textit{Attraction}}}\\ \hdashline
Input MR    & \textit{inform(area=north;price=no entrance fee; type=park; phone=01223)}\\
Reference   & yes, it's a park located in the north area of town and has no entrance fee. the phone number is 01223. \\
FT-GPT   & yes, there is \textcolor{blue}{a free entrance fee of £1}. phone 01223 for more information. \textit{\textbf{\textcolor{red}{\% miss \{area=north\}}}} \\
SC-GPT  & no problem. 01223 is the phone number and \textcolor{green}{the park is north}. the entrance fee is free.\\
\augnlg-FT & yes, \textcolor{green}{the entrance fee is no entrance fee}. the park is in the north. phone is 01223.\\
\augnlg-SC & yes, the park is in the north. no entrance fee. phone number 01223. \\
\hline
\end{tabular}
\caption{Example utterances generated by different models on \fewshotwoz{} (Better viewed in color). Errors are shown in three colors. The \textcolor{red}{red} text starting with ``\%'' denotes omission. The \textcolor{blue}{blue} text indicates hallucination. The \textcolor{green}{green} text means non-fluent generation. \label{table:examples_woz}}
\end{table*}

\paragraph{Qualitative Evaluation}
\autoref{table:examples_woz} compares some generated utterances by different models on \fewshotwoz{} (examples in \fewshotsgd{} are shown in \autoref{table:examples_sgd} in Appendix~\ref{sec:sup-example_sgd}). 
Both FT-GPT and SC-GPT are prone to omit important slots. 
Comparing to SC-GPT, FT-GPT tends to over-generate and introduces hallucinations. 
However, \augnlg{} and \augnlg{}-SC managed to generate fluent, natural text 
while precisely reflecting the the input MR.
We further examined 70 randomly sampled utterances generated by \augnlg{-SC}, whose BLEU scores are \textbf{lower} than those generated by SC-GPT, in the ``\textit{Hotel}'', ``\textit{Train}'' and ``\textit{Taxi}'' domain to understand some potential factors causing the lower BLEU scores 
We found that the lower BLEU scores are mainly driven by BLEU penalizing semantically correct paraphrases due to the nature of BLEU only checking surface-level matches. Some examples of such penalization are provided in \autoref{table:examples_woz_hotel} in Appendix~\ref{sec:sup-example_sgd}. Only 7 out of the 70 manually checked examples generated by \augnlg{-SC} are actually worse than SC-GPT.\footnote{We also examined 70 randomly sampled utterances generated by \augnlg{-SC}, whose BLEU scores are {\bf equal/higher} than those generated by SC-GPT. Among these examples, 35 examples are actually better and 7 examples are worse than the SC-GPT generations.}


In sum, the results (1) verify the effectiveness of complementing existing transfer learning methods with our novel data augmentation approach;
(2) reveal that automatically augmented MR-to-Text data alone can lead to a competitive performance, previously only achieved with existing MR-to-Text data. Since existing MR-to-Text data is not a scalable data source, our approach brings more practical values to real-world applications; 
(3) indicate that leveraging augmented MR-to-Text data on top of existing MR-to-Text data yields a new SOTA performance on the benchmark test.

\section{In-depth Analysis}
In this section, we provide comprehensive analysis results on the key components and parameters of our system to gain detailed insights: (1) intrinsic evaluation on augmented data, (2) influence of NLU quality, and (3) performance trends over varying amounts of augmented data. 


\begin{table}[t]
	\small\centering
	\begin{tabular}{l|p{0.3cm}p{0.3cm}p{0.3cm}p{0.3cm}p{0.3cm}p{0.3cm}p{0.3cm}}
	\hline
	{\bf Metrics} & {\bf Re} & {\bf La} & {\bf Ho} & {\bf TV} & {\bf At} & {\bf Tr} & {\bf Ta} \\ \hdashline
	{\bf MR Cov. $\uparrow$}  & .70 & .21 & .71 & .40 & .66 & .44 & .59 \\ 
	{\bf SL Cov. $\uparrow$}  & 1.0 & .95 & 1.0 & .94 & .89 & .92 & .86 \\ 
	\hline
	\end{tabular}
	\caption{Augmented data evaluation of \textit{MR Cov.} and \textit{SL Cov.} on \fewshotwoz{}. The domain names are represented by the first two letters. \label{table:augment_woz_mr_sl}}
\end{table}

\begin{table}[t]
	\small\centering
	\begin{tabular}{l|p{0.3cm}p{0.3cm}p{0.3cm}p{0.3cm}p{0.3cm}p{0.3cm}p{0.3cm}p{0.3cm}} \hline
	{\bf Metrics}        & {\bf Re} & {\bf Ho} & {\bf Fl} & {\bf Ca} & {\bf Ba} & {\bf We} & {\bf Bu} & {\bf Ev} \\ \hdashline
	{\bf MR Cov $\uparrow$}        & .80 & .72 & .66 & .65 & .43 & .70 & .58 & .80 \\ 
	{\bf SL Cov $\uparrow$}      & .92 & .85 & .89 & .75 & .57 & .86 & .88 & .93 \\
	\hline\hline
	{\bf Metrics}        & {\bf Ho} & {\bf Me} & {\bf Mo} & {\bf Mu} & {\bf Re} & {\bf Ri} & {\bf Se} & {\bf Tr} \\ \hdashline
	{\bf MR Cov $\uparrow$}        & .59 & .58 & .74 & .67 & .81 & .77 & .88 & .55 \\ 
	{\bf SL Cov $\uparrow$}      & .75 & .67 & .80 & .75 & .80 & .78 & .93 & .71 \\
	\hline
	\end{tabular}
	\caption{Augmented data evaluation of \textit{MR Cov.} and \textit{SL Cov.} on \fewshotsgd{}. The domain names are represented by the first two letters. \label{table:augment_sgd_mr_sl}}
\end{table}

\begin{table*}[t]
	\small\centering
	\begin{tabular}{c|l|ccccccc}
	\hline
	{\bf Metrics} & {\bf Data} & {\bf Restaurant} & {\bf Laptop} & {\bf Hotel} & {\bf TV} & {\bf Attraction} & {\bf Train} & {\bf Taxi} \\ \hline
	\multirow{2}{*}{\bf PPL $\downarrow$} & EXIST & 04.14 & 22.92 & 04.09 & 19.53 & 08.28 & 09.04 & 06.74 \\ 
	& AUG & {\bf 03.48} & {\bf 08.46} & {\bf 02.89} & {\bf 05.77} & {\bf 04.73} & {\bf 06.77} & {\bf 06.72} \\ \hdashline
	\multirow{2}{*}{\bf Nvt. (\%) $\downarrow$} & EXIST & 57.36 & 71.11 & 55.21 & 72.34 & 55.37 & {\bf 53.45} & {\bf 46.94} \\
	& AUG & {\bf 54.50} & {\bf 50.73} & {\bf 48.39} & {\bf 44.93} & {\bf 39.83} & 56.24 & 55.38 \\ 
	\hline
	\end{tabular}
	\caption{Language Model perplexity (PPL) and average n-gram novelty (Nvt.) on augmented data. \label{table:augment_nvt_ppl}}
\end{table*}

\begin{figure*}[tb]
	\centering
	\includegraphics[width=0.9\textwidth]{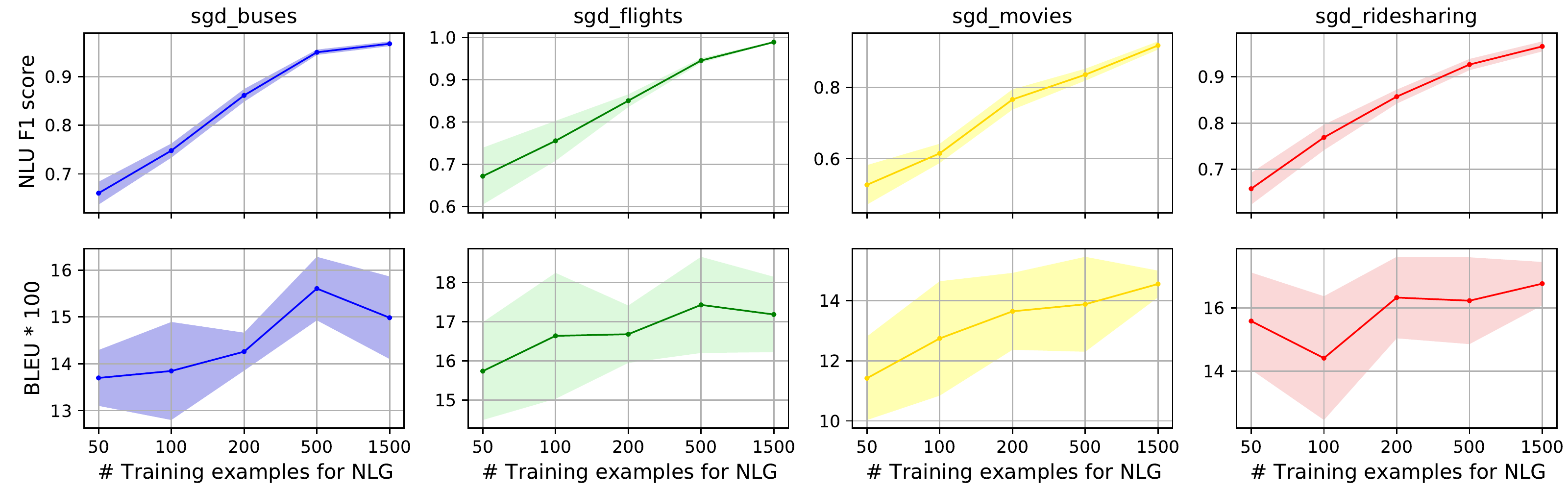}
	\caption{The influence of NLU on four domains in \fewshotsgd{}. The top row shows NLU F1 scores with \textit{50, 100, 200, 500, 1500} training examples. The bottom row shows the BLEU scores of \augnlg{}-FT pre-trained using these NLU models. All experiments are repeated for 5 times with different samples.}
	\label{fig:influence_nlu}
\end{figure*}

\subsection{Intrinsic Evaluation on Augmented Data} 

For intrinsic evaluation of augmented data, we first introduce four metrics:

$\bullet$ \textit{MR coverage (MR Cov.)}
evaluates the coverage of delexicalized MRs of the test set in the augmented set:
\begin{equation*}\label{Eq:5}
	\textup{MR Cov.} = \frac{\textup{\# delexicalized MRs} \in \mathbb{A}' \cap \mathbb{A}_{\textup{test}}}{\textup{\# delexicalized MRs} \in \mathbb{A}_{\textup{test}}},
\end{equation*}
where $\mathbb{A}'$ and $\mathbb{A}_{\textup{test}}$ denote delexicalized MRs in the augmented set and the test set, respectively. Higher \textit{MR Cov.} values indicate that more delexicalized MRs of the test set appear in the augmented set.

$\bullet$ \textit{Slot coverage (SL Cov.)}
evaluates the coverage of slot keys of the test set in the augmented set. 


$\bullet$ \textit{Language model perplexity (PPL)}
is the perplexity of augmented utterances calculated by a GPT-2 language model fine-tuned on the test set. Lower \textit{PPL} values indicate that the distribution of augmented utterances is close to that of the test utterances.

$\bullet$ \textit{Average n-gram novelty (Nvt.)} 
N-gram novelty measures the fraction of the n-grams in the test set that do not appear in the augmented set:
\begin{equation*}\label{Eq:6}
	\textup{N-gram novelty} = 1 - \frac{\textup{\# n-grams} \in \mathbb{X}' \cap \mathbb{X}_\textup{test}}{\textup{\# n-grams} \in \mathbb{X}_\textup{test}},
\end{equation*}
where $\mathbb{X}'$ and $\mathbb{X}_\textup{test}$ denote utterances in the augmented set and test set, respectively. Lower \textit{Nvt.} values indicate that more n-grams of the test set appear in the augmented set. 
We consider from 1-grams to 4-grams and report the average value.

The results of \textit{MR Cov.} / \textit{SL Cov.} on \fewshotwoz{} and \fewshotsgd{} are shown in \autoref{table:augment_woz_mr_sl} and \autoref{table:augment_sgd_mr_sl}, respectively. \textit{SL Cov.} achieves 70\% in most domains on both datasets while \textit{MR Cov.} has a wide range of values across domains. 
Noteworthily, \autoref{table:augment_sgd_mr_sl} strongly correlates with \autoref{table:fewshotsgd_res} -- ``Banks'' and ``Media'' domains are worse than other domains in both coverage metrics and NLG performance. On the other hand, ``Restaurants'' and ``Events'' domains are better than the others in both aspects. Although we do not see the same pattern on \fewshotwoz{}, it could be attributed to the large variance in the number of delexicalized MRs in each domain (see Table~2 in \cite{peng-etal-2020-shot}).

The results of \textit{PPL} and \textit{Nvt.} on {\fewshotwoz} are shown in \autoref{table:augment_nvt_ppl}. 
We compare the augmented data (\textit{AUG}) with the existing MR-to-Text data (\textit{EXIST}).
The top section shows that \textit{AUG} achieves lower \textit{PPL} values in all seven domains compared to \textit{EXIST}. 
The bottom section again demonstrates that \textit{AUG} achieves lower \textit{Nvt.} values in most domains.
However, in the ``\textit{Train}'' and ``\textit{Taxi}'' domains \textit{EXIST} attains lower novelty values, which matches the results in \autoref{table:fewshotwoz_res}, SC-GPT outperforming {\augnlg}-SC in these two domains.\footnote{
    Detailed breakdowns of novelty scores from 1-grams to 4-grams are provided in \autoref{table:augment_woz} in Appendix~\ref{sec:sup-augdata-statistic}.
	The \textit{Nvt.} results on \fewshotsgd{} are shown in \autoref{table:augment_sgd} in Appendix~\ref{sec:sup-augdata-statistic}, demonstrating similar trends.} 

\subsection{{Influence of NLU}}
\label{sub:discussion_nlu_influence}

\paragraph{Few-shot NLU performance}
Since few-shot NLU models are a key component of our system, we report their performance in F1 score. 
For each domain, we evaluate the few-shot NLU model on the Text-to-MR test set, prepared in Section~\ref{subsec:annotation}.
The average F1 over all domains on {\fewshotwoz} and {\fewshotsgd} are 0.77 and 0.68, respectively.
A further breakdown over the domains are provided in \autoref{table:nlu_woz} and \autoref{table:nlu_sgd} in Appendix~\ref{sec:sup-fewshot-nlu}. 

\paragraph{Influence of NLU Quality}
The mediocre NLU performance on \fewshotsgd{} leads to the following research question: \emph{can better NLU models boost NLG performance?}
To answer this question, we select four domains from \fewshotsgd{} with relatively low NLU performance: ``\textit{Buses} (0.63)'', ``\textit{Flights} (0.74)'', ``\textit{Movies} (0.44)'', and \textit{Ridesharing} (0.63). In each domain, we construct a new test set by randomly sampling 500 MR-to-Text pairs from the original test set, and take the rest as the NLU training pool.
To obtain NLU models of varying quality, we train a set of models while varying the amount of training data with stratified sampling.
The top row in \autoref{fig:influence_nlu} clearly shows that F1 score increases in proportion to the training size, reaching 0.95 in F1 in all four domains.
We then annotate the augmented utterances with different NLU models and pre-train the NLG models with the augmented MR-to-Text data updated with new MR labels.
Finally, we fine-tune the NLG models on the in-domain training set $\mathbb{D}$ and perform evaluation on the newly constructed 500 test set.
The bottom row in \autoref{fig:influence_nlu} confirms that there is a general proportional relationship between the performances of NLU and NLG.

\subsection{Varying Amounts of Augmentation} 

\begin{table}[t]
	\small\centering
	\begin{tabular}{l|c|ccccc}
	\hline
	{\bf Do}& {\bf Mod} & {\bf 50} & {\bf 100} & {\bf 200} & {\bf 500} & {\bf 1500} \\ \hline\hline
	\multirow{2}{*}{Bu} & {\bf FT} & 7.87 & 10.38 & 15.21 & 21.83 & 24.91 \\
	& {\bf AUG} & 14.37 & 15.36 & 17.06 & 22.18 & 24.98 \\ \hdashline
	\multirow{2}{*}{Fl} & {\bf FT} & 10.40 & 12.93 & 19.91 & 25.97 &  29.18 \\
	& {\bf AUG} & 14.07 & 15.50 & 21.55 & 25.38 & 26.62 \\ \hdashline
	\multirow{2}{*}{Mo} & {\bf FT} & 13.30 & 16.13 & 21.99 & 29.76 & 34.04 \\
	& {\bf AUG} & 17.13 & 17.55 & 23.68 & 29.14 & 33.55 \\ \hdashline
	\multirow{2}{*}{Ri} & {\bf FT} & 12.32 & 16.99 & 23.25 & 27.99 &  29.02 \\
	& {\bf AUG} & 17.18 & 22.06 & 24.76 & 26.87 & 28.60 \\
	\hline
	\end{tabular}
	\caption{BLEU scores for FT-GPT (FT) and \augnlg-FT (AUG) with different training sizes (50, 100, 200, 500, 1500). ``Bu'', ``Fl'', ``Mo'' and ``Ri'' are short for the domain names ``Buses'', ``Flights'', ``Movies'', ``Ridesharing''. All experiments are repeated for 5 times with different samples. \label{table:moredata}}
	\vspace{-12pt}
\end{table}
Lastly, we investigate the relationship between the amount of in-domain ground-truth data and the effect of augmentation. 
As in the previous section, we build new test sets by randomly taking 500 examples and vary the size of training set to train both NLU and NLG models.
\autoref{table:moredata} shows that, in all four domains, the performance difference between \augnlg{}-FT and FT-GPT culminates at the smallest training set and gradually diminishes as more training data become available. 

\section{Conclusion}
In this paper, we proposed {\augnlg}, a novel data  augmentation approach that combines a self-trained  retrieval model with a few-shot learned NLU, to automatically create MR-to-Text data from open-domain texts.
Experimental results verify the effectiveness of our approach by establishing new SOTA performances on two benchmark tests.
More importantly, we showed how our approach complements the previous SOTA approach, which hinges on unscalable data sources,  with unlimited open-domain data.  
Future work includes (1) technical innovations on each component of our system for further performance improvements, (2) exploring self-training on the NLU side too to evolve both the NLU and NLG model at the same time.

\section*{Acknowledgments}
We would like to thank the first author of \citet{peng-etal-2020-shot}, Baolin Peng, for his generous help. We also thank the anonymous reviewers for their helpful comments.

\bibliography{acl2021}
\bibliographystyle{acl_natbib}

\clearpage
\onecolumn
\appendix

\section{The calculation of TF-IDF} \label{sec:sup-tf-idf}
To calculate the TF-IDF score for a n-gram phrase, we take all in-domain texts $\mathbb{X}$ as one document $d$ to calculate its TF (Term Frequency) score, and randomly selected open-domain texts as the set of documents $D$ to calculate the IDF (Inverse Document Frequency) score\footnote{Here, each open-domain text represents a document.}. Thus, we formulate the TF-IDF score for n-gram phrase $ph_i$ as:
\begin{equation*}\label{Eq:3}
    \small
	\textup{TF-IDF} \left ( ph_i, d, D \right ) = \textup{tf}\left ( ph_i, d \right ) \cdot \textup{idf}\left ( ph_i, D \right ),
\end{equation*}
where,
\begin{equation*}\label{Eq:4}
    \small
	\begin{split}
		\textup{tf}\left ( ph_i, d \right ) &= \log \left ( 1+\textup{freq}\left ( ph_i, d \right ) \right ) \\
		\textup{idf}\left ( ph_i, D \right ) &= \log \left ( \frac{\left | D \right |}{ \left | \left \{ ph_i \in d \right \} \right | }  \right ),
	\end{split}
\end{equation*}
in which, $\textup{freq} \left ( ph_i, d \right )$ denotes the raw count of the phrase $ph_i$ appears in the document $d$.

\section{The structure of the BERT-based NLU annotation} \label{sec:sup-nlu}
\begin{figure}[!h]
	\centering
	\includegraphics[width=0.5\columnwidth]{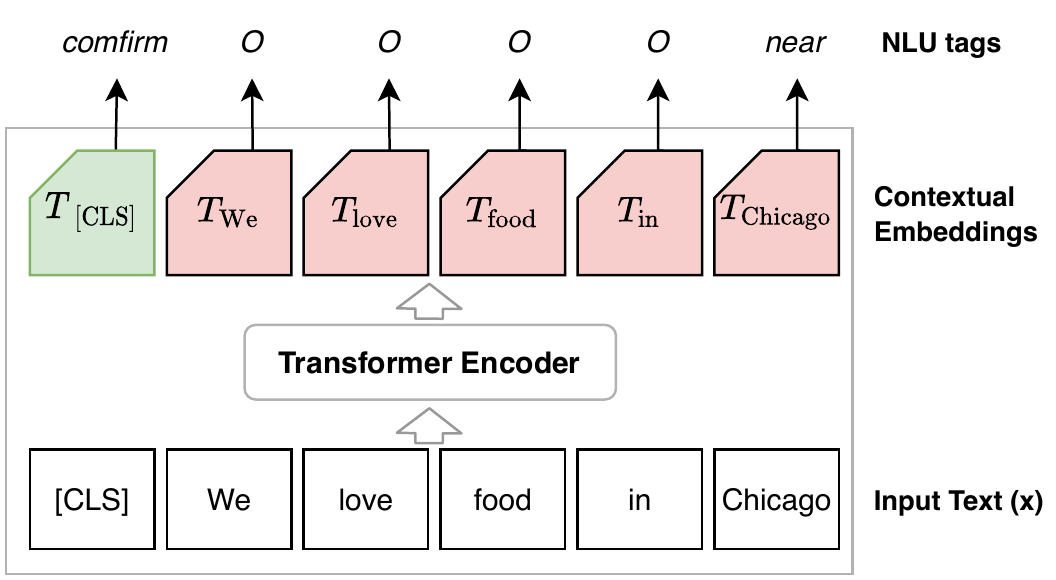}
	\caption{The structure of the BERT-based NLU annotation. The MR for the text ``We love food in Chicago'' is ``\textit{confirm ( near = Chicago )}''. Each slot-value token is annotated with the slot-name. The rest tokens are annotated with ``\textit{O}''.}
	\label{fig:nlu}
	\vspace{-12pt}
\end{figure}

\section{Statistics for the Augmented Data} \label{sec:sup-augdata-statistic}

\begin{table}[!h]
	\small\centering
	\begin{tabular}{l|ccccccc}
	\hline
	{\bf Domains}	& {\bf Restaurant} & {\bf Laptop} & {\bf Hotel} & {\bf TV} & {\bf Attraction} & {\bf Train} & {\bf Taxi} \\ \hdashline
	\# InD Pairs & 51 & 51 & 51 & 51 & 50 & 50 & 40\\
	\# Keywords (K)	& 0.23 & 0.06 & 0.22 & 0.28 & 0.27 & 0.18 & 0.17 \\ 
	\# Rtv Texts (K)	& 885.46 & 1000.13 & 760.61 & 850.00 & 1262.69 & 650.53 & 573.93 \\
	\# Aug Pairs (K)	& 30.97 & 36.62 & 40.46 & 49.76 & 65.48 & 9.60 & 4.95 \\ 
	\# Delex MRs (K) & 0.78 & 5.84 & 0.91 & 6.39 & 0.33 & 0.54 & 0.05\\
	\hline
	\end{tabular}
	\caption{{\fewshotwoz} statistics of the augmented pairs over 7 different domains. InD is short for in-domain. \label{table:augment_data_woz}}
	\vspace{-12pt}
\end{table}

\begin{table}[!tbh]
	\small\centering
	\begin{tabular}{l|cccccccc}
	\hline
	{\bf Domains}        & {\bf Restaurants} & {\bf Hotels} & {\bf Flights} & {\bf Calendar} & {\bf Banks} & {\bf Weather} & {\bf Buses} & {\bf Events} \\ \hdashline
	\# InD Pairs       & 50 & 50 & 50 & 25 & 23 & 11 & 50 & 50 \\ 
	\# Keywords (K)      & 0.23 & 0.15 & 0.25 & 0.09 & 0.05 & 0.04 & 0.17 & 0.17 \\
	\# Rtv Texts (K)      & 1021.64 & 1068.43 & 1195.41 & 582.22 & 112.78 & 387.90 & 749.46 & 1305.41 \\
	\# Aug Pairs (K)      & 61.51 & 20.64 & 39.59 & 56.87 & 1.27 & 6.39 & 11.15 & 56.55 \\
	\# Delex MRs (K) & 1.15 & 0.77 & 1.64 & 0.19 & 0.03 & 0.04 & 1.31 & 1.05 \\
	\hline\hline
	{\bf Model}        & {\bf Homes} & {\bf Media} & {\bf Movies} & {\bf Music} & {\bf Rentalcars} & {\bf Ridesharing} & {\bf Services} & {\bf Travel} \\ \hdashline
	\# InD Pairs       & 21 & 14 & 30 & 21 & 50 & 48 & 50 & 14 \\ 
	\# Keywords (K)      & 0.07 & 0.04 & 0.08 & 0.08 & 0.12 & 0.18 & 0.22 & 0.04 \\
	\# Rtv Texts (K)      & 403.91 & 335.42 & 538.68 & 1033.63 & 469.95 & 1180.02 & 953.51 & 362.45 \\
	\# Aug Pairs (K)      & 8.04 & 3.90 & 5.90 & 29.69 & 6.41 & 27.02 & 60.09 & 14.80 \\
	\# Delex MRs (K) & 0.15 & 0.05 & 0.06 & 0.13 & 0.20 & 0.23 & 2.00 & 0.05 \\
	\hline
	\end{tabular}
	\caption{{\fewshotsgd} statistics of the augmented pairs over 16 domains. InD is short for in-domain. \label{table:augment_data_sgd}}
	\vspace{-12pt}
\end{table}

\begin{table}[!h]
	\small\centering
	\begin{tabular}{l|cccccccc}
	\hline
	{\bf Nvt. (\%)} $\downarrow$	& {\bf Data} & {\bf Restaurant} & {\bf Laptop} & {\bf Hotel} & {\bf TV} & {\bf Attraction} & {\bf Train} & {\bf Taxi} \\ \hline\hline
	Nvt. uni & \multirow{4}{*}{\rotatebox{90}{{\bf EXIST}}} & 12.46 & 28.93 & 11.36 & 27.55 & 08.84 & 12.19 & 09.22 \\
	Nvt. bi && 48.70 & 69.82 & 46.68 & 72.53 & 46.66 & 46.13 & {\bf 35.40} \\
	Nvt. tri && 77.21 & 88.68 & 74.33 & 91.33 & 75.48 & {\bf 70.75} & {\bf 62.69} \\
	Nvt. four && {\bf 91.07} & 97.02 & {\bf 88.46} & 97.94 & 90.49 & {\bf 84.74} & {\bf 80.46} \\	
	\hdashline
	Nvt. uni & \multirow{4}{*}{\rotatebox{90}{{\bf AUG}}} & {\bf 11.20} & {\bf 06.33} & {\bf 06.13} & {\bf 04.51} & {\bf 04.09} & {\bf 09.80} & {\bf 07.56} \\
	Nvt. bi && {\bf 39.45} & {\bf 37.86} & {\bf 31.37} & {\bf 25.72} & {\bf 21.60} & {\bf 38.68} & 39.00 \\
	Nvt. tri && {\bf 73.63} & {\bf 69.10} & {\bf 66.98} & {\bf 61.10} & {\bf 53.72} & 80.21 & 79.10 \\
	Nvt. four && 93.73 & {\bf 89.63} & 89.10 & {\bf 88.39} & {\bf 79.92} & 96.28 & 95.87 \\
	\hline
	\end{tabular}
	\vspace{-2mm}
	\caption{N-gram novelty ($\downarrow$) breakdowns in {\fewshotwoz}. \label{table:augment_woz}}
	\vspace{-2mm}
\end{table}

\begin{table}[!h]
	\small\centering
	\begin{tabular}{l|ccccccccc}
	\hline
	{\bf Nvt. (\%) $\downarrow$}	& {\bf Data} & {\bf Restaurants} & {\bf Hotels} & {\bf Flights} & {\bf Calendar} & {\bf Banks} & {\bf Weather} & {\bf Buses} & {\bf Events} \\ \hdashline
	Nvt. uni	& \multirow{4}{*}{\rotatebox{90}{{\bf InD pairs}}} & 18.41 & 14.16 & 14.23 & 25.41 & 16.06 & 25.60 & 18.11 & 21.51 \\ 
	Nvt. bi	&& 58.08 & 53.84 & 59.32 & 69.45 & 49.91 & 72.73 & 66.16 & 62.22 \\
	Nvt. tri	 && 79.82 & 74.84 & 84.08 & 86.67 & 72.36 & 88.47 & 87.07 & 85.09 \\
	Nvt. four	&& 91.62 & {\bf 85.00} & 93.75 & 94.49 & {\bf 84.25} & 96.43 & 94.14 & 93.10 \\
	\hdashline
	Nvt. uni & \multirow{4}{*}{\rotatebox{90}{{\bf AUG}}} & {\bf 02.97} & {\bf 03.64} & {\bf 01.78} & {\bf 02.30} & {\bf 08.45} & {\bf 04.22} & {\bf 02.17} & {\bf 02.38} \\
	Nvt. bi && {\bf 18.94} & {\bf 21.63} & {\bf 19.54} & {\bf 17.22} & {\bf 37.91} & {\bf 34.61} & {\bf 28.09} & {\bf 19.88} \\
	Nvt. tri && {\bf 51.72} & {\bf 62.72} & {\bf 61.20} & {\bf 49.51} & {\bf 68.73} & {\bf 75.77} & {\bf 72.05} & {\bf 53.82} \\
	Nvt. four && {\bf 81.62} & 88.40 & {\bf 89.62} & {\bf 79.57} & 87.10 & {\bf 93.30} & {\bf 93.15} & {\bf 82.53} \\
	\hline\hline
	{\bf Nvt. (\%) $\downarrow$}	& {\bf Data} & {\bf Homes} & {\bf Media} & {\bf Movies} & {\bf Music} & {\bf Rentalcars} & {\bf Ridesharing} & {\bf Services} & {\bf Travel} \\ \hdashline
	Nvt. uni	& \multirow{4}{*}{\rotatebox{90}{{\bf InD pairs}}} & 30.73 & 30.85 & 32.06 & 35.91 & 19.81 & 15.71 & 21.59 & 42.11 \\ 
	Nvt. bi	&& 77.08 & 73.92 & 73.53 & 77.84 & 57.08 & 51.60 & 60.56 & 83.76 \\
	Nvt. tri	&& 90.78 & {\bf 88.90} & 87.20 & 91.35 & 76.91 & 78.27 & 81.44 & 93.68 \\
	Nvt. four	&& 95.22 & {\bf 93.68} & {\bf 92.77} & 96.55 & {\bf 87.09} & {\bf 89.74} & 92.37 & 97.59 \\
	\hdashline
	Nvt. uni	& \multirow{4}{*}{\rotatebox{90}{{\bf AUG}}} & {\bf 08.22} & {\bf 10.57} & {\bf 08.86} & {\bf 03.59} & {\bf 01.92} & {\bf 03.68} & {\bf 03.76} & {\bf 06.47} \\
	Nvt. bi	&& {\bf 37.85} & {\bf 60.10} & {\bf 47.53} & {\bf 32.38} & {\bf 26.68} & {\bf 25.48} & {\bf 24.82} & {\bf 38.29} \\
	Nvt. tri	&& {\bf 75.64} & 93.48 & {\bf 81.17} & {\bf 73.03} & {\bf 68.75} & {\bf 68.60} & {\bf 55.26} & {\bf 73.45} \\
	Nvt. four	&& {\bf 92.74} & 98.75 & 93.12 & {\bf 93.37} & 91.65 & 92.50 & {\bf 80.79} & {\bf 90.90} \\
	\hline
	\end{tabular}
	\vspace{-2mm}
	\caption{{\textit Nvt.} ($\downarrow$) breakdowns in {\fewshotsgd}. {\textit EXIST} are from the SGD, we compare with in-domain pairs. \label{table:augment_sgd}}
	\vspace{-6mm}
\end{table}

\section{Few-shot NLU Performance} \label{sec:sup-fewshot-nlu}
\vspace{-2mm}
\begin{table}[H]
	\small\centering
	\vspace{-2mm}
	\begin{tabular}{l|ccccccc}
	\hline
	{\bf Metrics}	& {\bf Restaurant} & {\bf Laptop} & {\bf Hotel} & {\bf TV} & {\bf Attraction} & {\bf Train} & {\bf Taxi} \\ \hdashline
	Precision	& 1.000 & .8229 & .7500 & .7904 & .6050 & .6552 & .6178 \\
	Recall	& 1.000 & .8490 & .7500 & .8382 & .6904 & .6706 & .7239 \\ 
	F1 score	& 1.000 & .8357 & .7500 & .8136 & .6449 & .6628 & .6667 \\
	\hline
	\end{tabular}
	\vspace{-2mm}
	\caption{NLU evaluation for {\fewshotwoz} (Precision$\uparrow$), (Recall$\uparrow$), (F1 score$\uparrow$) \label{table:nlu_woz}}
	\vspace{-6mm}
\end{table}

\begin{table}[H]
	\small\centering
	\begin{tabular}{l|cccccccc}
	\hline
	{\bf Metrics}        & {\bf Restaurants} & {\bf Hotels} & {\bf Flights} & {\bf Calendar} & {\bf Banks} & {\bf Weather} & {\bf Buses} & {\bf Events} \\ \hdashline
	Precision        & .6346 & .6516 & .7229 & .8332 & .8971 & .7177 & .6289 & .5333 \\ 
	Recall      & .6635 & .6866 & .7560 & .8897 & .8684 & .7183 & .6372 & .5870 \\
	F1 score      & .6487 & .6686 & .7391 & .8605 & .8825 & .7180 & .6330 & .5589 \\
	\hline\hline
	{\bf Metrics}        & {\bf Homes} & {\bf Media} & {\bf Movies} & {\bf Music} & {\bf Rentalcars} & {\bf Ridesharing} & {\bf Services} & {\bf Travel} \\ \hdashline
	Precision        & .8201 & .6404 & .4787 & .8011 & .5183 & .6145 & .5470 & .8132 \\ 
	Recall      & .8166 & .5060 & .3996 & .7889 & .5288 & .6546 & .6146 & .8713 \\
	F1 score      & .8183 & .5653 & .4356 & .7950 & .5235 & .6340 & .5788 & .8412 \\
	\hline
	\end{tabular}
	\vspace{-2mm}
	\caption{NLU evaluation for {\fewshotsgd} (Precision$\uparrow$), (Recall$\uparrow$), (F1 score$\uparrow$).\label{table:nlu_sgd}}
	\vspace{-5mm}
\end{table}

\section{Generation Examples} \label{sec:sup-example_sgd}
\vspace{-3mm}

\begin{table}[!tbh]
\centering\small
\vspace{-2mm}
\begin{tabular}{l|l}
\hline
\multicolumn{2}{c}{\textbf{Domain: \textit{Hotel }}} \\ \hdashline
Input MR    & \textit{inform(name=parker guest house; area=mastro; address=520 church street; phone=4156213222)}\\
Reference   & parker guest house is near mastro at 520 church street. their phone number is 4156213222. \\
SC-GPT  & the parker guest house is located in mastro and is at 520 church street. the phone number is 4156213222.\\
\augnlg-SC & the parker guest house in mastro's address is 520 church street. the phone number is 4156213222.\\
\hline

\multicolumn{2}{c}{\textbf{Domain:\textit{~Train}}}\\ \hdashline
Input MR    & \textit{inform(day=sunday; id=tr5413; arriveby=00:04 ) @ book (none)}\\
Reference   & train tr5413 arrives at 24:44 or 00:04 on sunday. would you like to book it now ? \\
SC-GPT  & tr5413 arrives at 00:04. would you like me to book it for you ? \\
\augnlg-SC & tr5413 arrives at 00:04 on sunday. would you like to book ?\\
\hline

\multicolumn{2}{c}{\textbf{Domain:~\textit{Taxi}}}\\ \hdashline
Input MR    & \textit{inform(arriveby=16:00)}\\
Reference   & yes , the taxi is due to arrive at 16:00. \\
SC-GPT  & yes , the taxi will arrive by 16:00.\\
\augnlg-SC & i 'm on it , the taxi will arrive by 16:00. \\
\hline
\end{tabular}
\vspace{-2mm}
\caption{Generation examples in ``\textit{Hotel}'', ``\textit{Train}'' and ``\textit{Taxi}'' domain on \fewshotwoz{}. In the same example, the BLEU scores for the SC-GPT generations are higher than the \augnlg-SC generations. \label{table:examples_woz_hotel}
\vspace{-2mm}
}
\end{table}

\begin{table}[!tbh]
\centering\small
\begin{tabular}{l|l}
\hline
\multicolumn{2}{c}{\textbf{Domain: \textit{Banks}}} \\ \hdashline
Input MR	& \textit{OFFER(balance=2,890.54;account\_type=checking)} \\
Reference	& your checking balance is 2,890.54 . \\
FT-GPT	& now you have 2,890.54 to spend on goods and services . \\
\augnlg-FT	& balance of checking account is 2,890.54 \\
\hline

\multicolumn{2}{c}{\textbf{Domain:\textit{ Buses }}}\\ \hdashline
Input MR	& \textit{OFFER(departure\_time=10:50 am;price=42;fare\_type=economy)} \\
Reference	& there is a bus leaving at 10:50 am . its economy ticket costs 42 \\
FT-GPT	& please confirm that you want to leave from 10:50 am to 10:50 pm . \\
\augnlg-FT	& sure , when do you want the bus to leave at ? 10:50 am , economy ticket or 42 ? \\
\hline

\multicolumn{2}{c}{\textbf{Domain:\textit{Calendar}}}\\ \hdashline
Input MR	& \textit{OFFER(available\_start\_time=8 am;available\_end\_time=7:30 pm ) @ INFORM\_COUNT (count)} \\
Reference	& yes , you have 2 empty slots from 8 am to 7:30 pm . \\
FT-GPT	& you have 1 available seat . \\
\augnlg-FT	& is your event scheduled to start at 8 am and end at 7:30 pm ? \\
\hline

\multicolumn{2}{c}{\textbf{Domain:~\textit{Events}}}\\ \hdashline
Input MR	& \textit{CONFIRM(number\_of\_tickets=1;city=new york)} \\
Reference	& no problem , i 'll make it 1 ticket for new york . \\
FT-GPT	& please confirm your number of tickets . \\
\augnlg-FT	& 1 ticket to the event in new york . \\
\hline

\multicolumn{2}{c}{\textbf{Domain:~\textit{Flights}}}\\ \hdashline
Input MR	& \textit{CONFIRM(airlines=southwest airlines;return\_date=march 12th)} \\
Reference	& so you want a southwest airlines flight that returns on march 12th ? \\
FT-GPT	& it arrives on march 12th at southwest airlines . \\
\augnlg-FT	& it's not an upgrade , it's a southwest airlines flight that leaves on march 12th . \\
\hline

\multicolumn{2}{c}{\textbf{Domain:~\textit{Homes}}}\\ \hdashline
Input MR	& \textit{INFORM(phone\_number=650-961-2975;furnished=false)} \\
Reference	& the phone number is 650-961-2975 . and the apartment is not furnished . \\
FT-GPT	& you are willing to call 650-961-2975 to schedule an appointment . \\
\augnlg-FT	& 650-971-2975 is the phone number for the furnished apartment . \\
\hline

\multicolumn{2}{c}{\textbf{Domain:~\textit{Movies}}}\\ \hdashline
Input MR	& \textit{OFFER(movie\_name=hackers, high life, sunset)} \\
Reference	& i found hackers , high life , and sunset for you \\
FT-GPT	& what movie would you like ? \\
\augnlg-FT	& what movies are you interested in seeing then ? \\
\hline

\multicolumn{2}{c}{\textbf{Domain:~\textit{Music}}}\\ \hdashline
Input MR	& \textit{CONFIRM(song\_name=mile high;playback\_device=bedroom speaker)} \\
Reference	& just to make sure , you want to listen to mile high on the bedroom speaker . \\
FT-GPT	& you are listening to the song mile high by mile high . \\
\augnlg-FT	& what is your favorite song from the mile high club ? ( or any song for that matter ) \\
\hline

\multicolumn{2}{c}{\textbf{Domain:~\textit{Rentalcars}}}\\ \hdashline
Input MR	& \textit{REQUEST(pickup\_date=march 5th;dropoff\_date=march 9th)} \\
Reference	& would you like to rent it on march 5th and return it on march 9th ? \\
FT-GPT	& what date would you like the car to arrive from ? \\
\augnlg-FT	& will you be picking it up on march 5th and returning it on march 9th ? \\
\hline

\multicolumn{2}{c}{\textbf{Domain:~\textit{Ridesharing}}}\\ \hdashline
Input MR	& \textit{INFORM(wait\_time=2;ride\_fare=36.01)} \\
Reference	& the cab will arrive in 2 minutes and it's going to cost 36.01 . \\
FT-GPT	& what is the fare for 2 ? \\
\augnlg-FT	& please wait 2 hours and 36.01 will be added to your cart . \\
\hline

\multicolumn{2}{c}{\textbf{Domain:~\textit{Services}}}\\ \hdashline
Input MR	& \textit{INFORM(is\_unisex=true;phone\_number=925-446-4144 ) @ NOTIFY\_SUCCESS ( none)} \\
Reference	& it is uniesex and you have an appointment set . phone them at 925-446-4144 . \\
FT-GPT	& your appointment has been scheduled . the salon is unisex and phone is 925-446-4144 . \\
\augnlg-FT	& your appointment is confirmed . the unisex bathroom is located at 925-446-4144 . \\
\hline

\multicolumn{2}{c}{\textbf{Domain:~\textit{Weather}}}\\ \hdashline
Input MR	& \textit{INFORM(humidity=43)} \\
Reference	& it will be around 43 \% . \\
FT-GPT	& the humidity will be 43 percent . \\
\augnlg-FT	& the humidity is 43 percent . \\
\hline

\multicolumn{2}{c}{\textbf{Domain:~\textit{Media}}}\\ \hdashline
Input MR	& \textit{INFORM(directed\_by=roxann dawson ) @ NOTIFY\_SUCCESS ( none)} \\
Reference	& okay , i started the movie . the director is roxann dawson . \\
FT-GPT	& the movie has started . \\
\augnlg-FT	& it's directed by roxann dawson and it's calledn'tify success . \\
\hline

\multicolumn{2}{c}{\textbf{Domain:~\textit{Travel}}}\\ \hdashline
Input MR	& \textit{INFORM(phone\_number=416-646-4677;good\_for\_kids=true)} \\
Reference	& the phone number is 416-646-4677 and is a nice place for kids . \\
FT-GPT	& yes , it is a good place for kid -s . \\
\augnlg-FT	& the phone number is 416-646-4677. yes it's a nice place for kid -s . \\
\hline
\end{tabular}
\caption{Randomly sampled generation examples from {\fewshotsgd}.\label{table:examples_sgd}}
\end{table}

\end{document}